\documentclass{article}

\usepackage[preprint]{neurips_2025}  
\usepackage[utf8]{inputenc} 
\usepackage[T1]{fontenc}    
\usepackage{hyperref}       
\usepackage{url}           
\usepackage{booktabs}       
\usepackage{amsmath,amssymb,amsthm}
\usepackage{amsfonts}
\usepackage{nicefrac}       
\usepackage{microtype}      
\usepackage{xcolor}         
\usepackage{graphicx}
\usepackage{subcaption}
\usepackage{enumitem}
\usepackage{algorithm}
\usepackage{algorithmic}
\usepackage{multirow}
\usepackage{tabularx}

\newtheorem{theorem}{Theorem}[section]

\theoremstyle{remark}

\newcommand{\E}{\mathbb{E}}
\newcommand{\R}{\mathbb{R}}
\newcommand{\Var}{\mathrm{Var}}

\title{Targeted Deep Architectures: A TMLE-Based Framework for Robust Causal Inference in Neural Networks}

\author{Yi Li, David Mccoy, Nolan Gunter, Kaitlyn Lee, Alejandro Schuler, Mark van der Laan}

\begin{document}

\maketitle


\begin{abstract}
Modern deep neural networks are powerful predictive tools yet often lack valid inference for causal parameters, such as treatment effects or entire survival curves. While frameworks like Double Machine Learning (DML) and Targeted Maximum Likelihood Estimation (TMLE) can debias machine-learning fits, existing neural implementations either rely on “targeted losses” that do not guarantee solving the efficient-influence-function equation or computationally expensive post-hoc “fluctuations” for multi-parameter settings. We propose \emph{Targeted Deep Architectures} (TDA), a new framework that embeds TMLE \emph{directly} into the network’s parameter space with no restrictions on the backbone architecture. Specifically, TDA partitions model parameters—freezing all but a small “targeting” subset—and iteratively updates them along a \emph{targeting gradient}, derived from projecting the influence-functions onto the span of the gradients of the loss with respect to weights. This procedure yields plug-in estimates that remove first-order bias and produce asymptotically valid confidence intervals. Crucially, TDA easily extends to \emph{multi-dimensional} causal estimands (e.g., entire survival curves) by merging separate targeting gradients into a single universal targeting update. Theoretically, TDA inherits classical TMLE properties, including double robustness and semiparametric efficiency. Empirically, on the benchmark IHDP dataset (average treatment effects) and simulated survival data with informative censoring, TDA reduces bias and improves coverage relative to both standard neural-network estimators and prior post-hoc approaches. In doing so, TDA establishes a direct, scalable pathway toward rigorous causal inference within modern deep architectures for complex multi-parameter targets.

\end{abstract}

\section{Introduction}
\label{sec:intro}

In many fields---from healthcare policy to economics and environmental monitoring---the goal is often not just to \emph{predict} outcomes from data but to \emph{infer} causal or semiparametric parameters (e.g., average treatment effects, conditional survival probabilities). Such inference demands accurate \emph{uncertainty quantification}, namely valid confidence intervals or hypothesis tests for the parameter of interest. Modern machine learning (ML) methods, in particular deep neural networks, excel at prediction but do not intrinsically provide valid statistical inference; indeed, training exclusively for predictive performance (e.g., minimizing mean-squared error or cross-entropy) often yields biased estimators of causal quantities with unreliable uncertainty estimates~\cite{chernozhukov2018double,shalit2017estimating}. 

\paragraph{Causal Inference and TMLE.}
Semiparametric approaches such as Targeted Maximum Likelihood Estimation (TMLE)~\cite{van2011targeted,van2018targeted} address this challenge by constructing a low-dimensional ``$\varepsilon$-submodel'' around an initial ML estimator and \emph{solving} the efficient influence function (EIF) equation to debias it. This procedure yields attractive theoretical guarantees, including \emph{double robustness} (consistency if at least one nuisance component is correctly estimated) and \emph{semiparametric efficiency} (minimal possible variance under broad conditions). When the initial estimator is a black-box model, however, implementing TMLE can become cumbersome for high dimensional target parameters. In the context of deep neural networks, two primary strategies have emerged:  
\begin{itemize}[leftmargin=2em]
    \item \textbf{Targeted regularization}, in which an additional ``targeting loss'' is introduced during training~\cite{shi2019adapting,chernozhukov_riesznet_2022}. Although one may theoretically remove first-order bias by optimizing over this extended objective, in practice there is no guarantee that at convergence the efficient influence function is truly solved with respect to the critical ``perturbation parameter'' embedded in the loss. Bias may thus remain.
    \item \textbf{Post-hoc TMLE fluctuations}, which fix a trained network and append a small, low-dimensional parametric model to solve the EIF equation for a particular parameter~\cite{chernozhukov_riesznet_2022,shirakawa2024longitudinal}. While this procedure does remove first-order bias for single-parameter targets (e.g., an average treatment effect), it can become cumbersome in multi-parameter settings like survival curve estimation~\cite{rytgaard2024one}.
\end{itemize}

Indeed, estimating \emph{multiple} parameters (e.g., survival probabilities at each time point) would require either multiple targeted-loss functions (whose interplay is rarely well understood) or a large post-hoc update that complicates implementation and can be memory-intensive~\cite{rytgaard2024one}. Universal least-favorable submodels~\cite{van2018targeted} can, in principle, target all parameters at once but often demand complex analytic forms (e.g., integrating hazards repeatedly), making them unwieldy to implement.

\paragraph{Related work.}
\emph{Adaptive Debiased Machine Learning} (ADML)~\cite{van2023adaptive} demonstrates that data‐driven sub-model selection, when coupled with classical debiasing (one-step, TMLE, or sieve), can produce efficient—and in some cases superefficient—estimators. The autoDML framework~\cite{van2025automatic} is a specific instance of ADML that elegantly handles smooth functionals of M-estimands through the Hessian Riesz representer found by minimizing $E_0[\ddot{\ell}_{\eta_0}(\theta_0)(\alpha, \alpha) - 2\dot{m}_{\theta_0}(\alpha)]$.

However, autoDML's reliance on Hessian operators limits its applicability in certain settings. When the Hessian operator $\ddot{\ell}_{\eta}(\theta)$ satisfies well-conditioning assumptions (Condition C5: positive definiteness with uniform lower bound), autoDML provides an elegant approach for smooth functionals of M-estimands. In overparameterized neural networks where these conditions often fail due to singular or near-singular Hessians, and where computing Hessian operators is computationally expensive and requires specialized implementation beyond standard automatic differentiation, TDA's gradient-only approach extends ADML's benefits to these challenging settings.

TDA represents an alternative ADML instance that overcomes both limitations. By projecting influence functions onto neural network gradients $\nabla_{\theta}\ell$ with regularization, TDA extends the theoretical applicability of ADML to settings where Hessian-based approaches fail, while remaining computationally tractable. Specifically, TDA: (i) handles overparameterized neural networks where autoDML's well-conditioning assumption is violated, (ii) works for any pathwise differentiable parameter beyond M-estimand functionals, (iii) accepts any valid influence function (IPTW, AIPW, or even autoDML's constructed EIF when available), and (iv) integrates seamlessly with modern automatic differentiation frameworks that efficiently compute gradients but struggle with second derivatives. This gradient-only approach brings rigorous statistical inference to modern deep architectures where traditional second-order methods are either theoretically inapplicable or computationally prohibitive.

Concurrently, \textbf{regHAL}~\cite{li2025regularized} develops regularized 
updates for HAL-induced working models using projection-based targeting 
that projects influence functions onto HAL score spaces with lasso/ridge 
penalties. TDA extends this projection-based philosophy to generic neural 
architectures, replacing HAL basis functions with neural network gradients 
and enabling application to any deep learning model. While regHAL focuses 
on stability within HAL's specific structure, TDA demonstrates that similar 
projection-based regularization principles apply broadly across modern 
deep architectures.

\begin{figure}[t]
\centering
\includegraphics[width=0.9\linewidth]{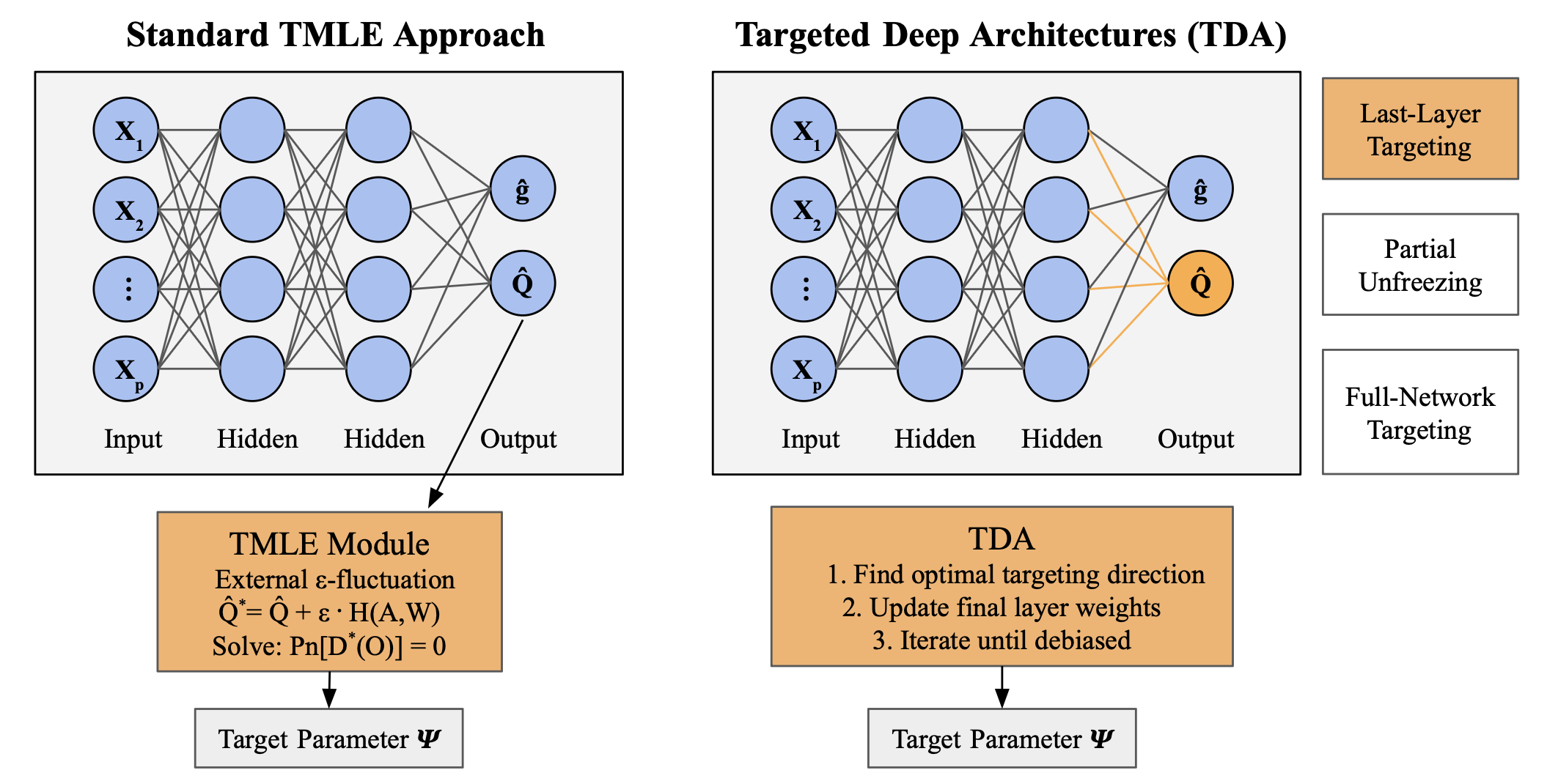}
\caption{\small \textbf{Traditional Post-Processing vs.\ Our Approach.} Left: conventional practice appends a small parametric “fluctuation” model outside a frozen neural net, which is cumbersome in multi-parameter settings. Right: TDA embeds the targeting step into a local final-layer parametric submodel, allowing universal updates for all parameters of interest.}
\label{fig:comparison}
\end{figure}

\paragraph{Our Contribution:}
We propose \emph{Targeted Deep Architectures (TDA)}, a framework that \emph{embeds} the TMLE update directly into a neural network’s parameter space.  Rather than adding an external fluctuation model (Figure~\ref{fig:comparison}, left), TDA \emph{freezes} most (or all) network weights and defines a small, local parametric fluctuation submodel using a “targeting” subset of parameters of neural network (often in the final layer).  By \emph{projecting} the relevant influence functions onto the partial derivatives of neural network loss wrt these parameters, TDA identifies a \emph{targeting gradient} that updates the fit to remove first-order bias.  Crucially, TDA can combine multiple targeting gradients into a single universal update, thereby targeting an entire vector of parameters (e.g., an entire survival curve) at once.  In this way, TDA provides a direct route to achieve valid statistical inference (i.e., unbiased point estimation and asymptotically correct confidence intervals) using standard deep learning architectures, but without complicated new losses or large post-hoc models.

\paragraph{Outline of Contributions and Paper Organization.}
\begin{enumerate}[leftmargin=1.25em,label=(\arabic*)]
  \item \textbf{We formulate TDA for neural networks} (Section~\ref{sec:method}), showing how to define a finite-dimensional “targeting submodel” in weight space and use \emph{influence-function projections} to derive a data-adaptive targeting gradient.
  \item \textbf{We show TDA generalizes to multiple parameters simultaneously} (Section~\ref{sec:method:multi}), including the important case of survival curve estimation under right censoring.
  \item \textbf{We prove TDA inherits classical TMLE guarantees} (Section~\ref{sec:theory}), leveraging recent results on adaptive debiased machine learning~\cite{van2023adaptive}: TDA achieves double robustness, semiparametric efficiency, and valid $\sqrt{n}$-inference, provided the submodel is large enough and appropriately selected.
  \item \textbf{We illustrate TDA empirically on (i) average treatment effect estimation} (Section~\ref{sec:experiments}), showing comparable or improved performance vs. existing single-parameter targeting approaches, and \textbf{(ii) survival analysis} under informative censoring, demonstrating TDA’s ease in simultaneously targeting entire survival curves.
\end{enumerate}

Hence, TDA paves a direct, scalable way to bring rigorous inference into deep neural models for causal and semiparametric parameters of interest.  Figure~\ref{fig:comparison} (right) illustrates the key contrast with standard post-hoc TMLE approaches.  

\section{Methodology: Targeted Deep Architectures}
\label{sec:method}

We consider i.i.d.\ observations $\{O_i\}_{i=1}^n \sim P_0$, where $P_0\in \mathcal{M}$ is unknown. Let $P_n$ denote the empirical distribution. Our goal is to estimate a (potentially vector-valued) parameter $\Psi(P_0)$ that is \emph{pathwise differentiable}; such parameters admit an \emph{efficient influence function} (EIF) $D^*_P$ at any $P\in\mathcal{M}$.  This EIF captures the optimal first-order correction for reducing bias in estimates of $\Psi(P_0)$ within $\mathcal{M}$~\cite{van2011targeted,van2018targeted}.  But in TDA, if $D^*_P$ is too complicated or even doesn't exist such as in survival analysis studies\cite{laan2003unified}, we don't need it but use an (unentered) influence function such as the one from the IPCW/IPTW estimator.

\paragraph{Example: Average Treatment Effect (ATE).}
For a classic causal inference problem, $O=(X,A,Y)\sim P_0\in \mathcal{M}_{np}$ with binary treatment $A\in\{0,1\}$, covariates $X$, and outcome $Y$.  The estimand
\[
  \Psi(P_0)\;=\; 
  \E_X\Bigl[\,\E\{Y\mid A=1,X\} \;-\;\E\{Y\mid A=0,X\}\Bigr]
\]
has an EIF at $P$ of
\[
  D^*_P(O)
  \;=\;
  \Bigl(\tfrac{A}{g(X)}-\tfrac{1-A}{1-g(X)}\Bigr)\bigl[Y - Q(A,X)\bigr]
  \;+\;
  \big(Q(1,X)-Q(0,X)\;-\;\Psi(P)\big),
\]
where $Q(a,x)=\E_P[Y\mid A=a,X=x]$ and $g(x)=P(A=1\mid X=x)$.

\paragraph{Initial Neural-Network Fit and Naive Plug-In.}
To model nuisance functions (e.g.\ $Q(a,x)$) flexibly, we train a neural network $f_{\theta}:\,\mathcal{X}\to\mathcal{Y}$ on part of the data.  Denote the resulting weights by $\hat{\theta}$.  For ATE, a naive \emph{plug-in} estimator is
\[
  \Psi\bigl(f_{\hat{\theta}}\bigr)
  \;=\; P_n\!\bigl[\hat{Q}(1,X)-\hat{Q}(0,X)\bigr],
  \quad
  \text{where }\;\hat{Q}(a,x)=f_{\hat{\theta}}(a,x).
\]
Because the network was trained primarily to minimize a predictive loss (e.g.\ MSE or negative log-likelihood) with heavy regularization, it does not necessarily remove the first-order bias for $\Psi$. In particular, naive plug-in estimates typically lack valid $\sqrt{n}$-confidence intervals.

\paragraph{Overview of Targeted Deep Architectures (TDA).}
To remove this bias without discarding the neural architecture, we embed the \emph{TMLE update} directly into the weight space.  Conceptually:

\begin{enumerate}[leftmargin=1.5em,label=(\arabic*)]
  \item \textbf{Partition the network weights} 
    $\theta = (\theta_{\text{fix}},\theta_{\text{targ}})$, freezing $\theta_{\text{fix}}$ and letting $\theta_{\text{targ}}$ vary.  This defines a parametric \emph{submodel} $M_{\text{targ}} = \{\,P_{\theta_{\text{fix}},\theta_{\text{targ}}}\colon \theta_{\text{targ}}\in\Theta_{\text{targ}}\}\subseteq\mathcal{M}$.
  \item \textbf{Approximate the EIF in Submodel}: Instead of trying to invert the empirical Fisher information (often ill-conditioned in overparameterized nets), we \emph{project} a consistent influence-function estimate onto the gradients $\nabla_{\theta_{\text{targ}}}\ell(\theta,O)$, with $\ell$ the network’s training loss.  This yields a \emph{targeting gradient} in weight space.
  \item \textbf{Iteratively update} the submodel parameter $\theta_{\text{targ}}$ along that targeting gradient until the empirical mean of the projected EIF is near zero, thus debiasing the plug-in estimate.
\end{enumerate}

Below, we detail these steps, culminating in a multi-parameter extension that simultaneously corrects many causal estimands.

\subsection{Defining a Parametric Submodel in Weight Space}
\label{sec:method:param_submodel}

Let $f_{\theta}(x)$ be our neural net with parameters $\theta\in\Theta$.  We define a small parametric \emph{submodel} by splitting
\[
  \theta \;=\; (\theta_{\text{fix}},\,\theta_{\text{targ}}),
\]
where $\theta_{\text{fix}}$ is held at its trained value and $\theta_{\text{targ}} \in \R^k$ is free to vary in a $k$-dimensional neighborhood.  For instance, $\theta_{\text{fix}}$ might include all early-layer weights, and $\theta_{\text{targ}}$ just the final layer. Formally, we define
$M_{\text{targ}}
  \;=\;
  \bigl\{
    P_{\theta_{\text{fix}},\,\theta_{\text{targ}}}
    \;\big|\;
    \theta_{\text{targ}} \in \Theta_{\text{targ}} \subseteq \R^k
  \bigr\}\subseteq \mathcal{M}$ and let $\Psi\bigl(P_{\theta_{\text{fix}},\,\theta_{\text{targ}}}\bigr)$ be the induced functional of interest. Usually, $\Psi\bigl(P_{\theta_{\text{fix}},\,\theta_{\text{targ}}}\bigr)\;=\;
\Psi\bigl(f_{\theta_{\text{fix}},\,\theta_{\text{targ}}}\bigr)$. Then, we choose $\theta_{\text{targ}}$ so that the empirical mean of the EIF at $P_{\theta_{\text{fix}},\theta_{\text{targ}}}$ wrt $\mathcal{M}$ goes to zero, i.e.\ $P_n\bigl[D^*_{P_{\theta_{\text{fix}},\theta_{\text{targ}}}}(O)\bigr]\approx0$, removing first-order bias.

\emph{Remark } Because we only require access to $\nabla_{\theta_{\text{targ}}}\ell\bigl(\theta,O_i\bigr)$, the procedure applies verbatim to CNNs, RNNs, attention blocks, etc.; the targeting subset can be any slice of the parameter vector.

\paragraph{ATE Submodel Example.}
One might let $X\sim P_{X,n}$ (the empirical distribution of $X$) and define 
\[
  Y\mid A,X \;\sim\; \mathcal{N}\!\bigl(f_{\theta_{\text{fix}},\theta_{\text{targ}}}(A,X),\;\sigma^2\bigr),\quad
  A\mid X\;\sim\;g(X),
\]
thereby capturing how $Y$ depends on $\theta_{\text{targ}}$ through the neural net’s final layer. The corresponding $\Psi\bigl(f_{\theta_{\text{fix}},\theta_{\text{targ}}}\bigr)$ is $E_{P_{X,n}}(f_{\theta_{\text{fix}},\theta_{\text{targ}}}(1,X)-f_{\theta_{\text{fix}},\theta_{\text{targ}}}(0,X))$.

\subsection{Parametric Submodel EIF and Targeting Gradient}
\label{sec:method:proj}

\paragraph{Local EIF in the Submodel.}
Classical TMLE would require the EIF of $\Psi$ wrt $\mathcal{M}$ to do "full" targeting since this EIF represents the direction in the function space $\mathcal{M}$ that provides the most efficient estimation for $\Psi(P_0)$. Nevertheless, if we want to stay in the parametric working submodel, we can define our working target estimand as the original target estimand restricted to $M_{\text{targ}}$ as $\Psi:M_{\text{targ}} \to R$ and our target parameter as $\Psi(P_{0,\textit{targ}})$ where $P_{0,\textit{targ}}=\arg\min_{P'\in M_{\text{targ}}}D_{KL}(P'||P_0)$.

Then, instead of doing "full" targeting in $\mathcal{M}$ for $\Psi(P_0)$, we can now perform "partial" targeting in $M_{\text{targ}}$ for $\Psi(P_{0,\textit{targ}})$. As we will prove in section \ref{sec:theory}, with careful choice of the submodel (large enough but not too data adaptively chosen), the later targeting also yields asymptotic efficient and regular estimates for $\Psi(P_0)$ in $\mathcal{M}$.

Let's denote $D^{*,\text{para}}_{\theta_{\text{targ}}}$ as the EIF for $\Psi$ at $\theta_{\text{targ}}$ wrt $M_{\text{targ}}$. Then as shown in \cite[Ch.~8]{van2018targeted} \cite{van2023adaptive}, $D^{*,\text{para}}_{\theta_{\text{targ}}}=\alpha^{*,T}_{\theta_{\text{targ}}} \nabla_{\theta_{\text{targ}}}\ell\bigl(\theta,O_i\bigr)$, where $ \alpha^*_{\theta_{\text{targ}}}=\arg\min_{\alpha} P_{\theta_{\text{fix}},\,\theta_{\text{targ}}}(D^*_{P_{\theta_{\text{fix}},\,\theta_{\text{targ}}}}-\alpha^\top \nabla_{\theta_{\text{targ}}}\ell\bigl(\theta,O_i\bigr))^2$. So that we are really doing "partial" targeting for the $\Psi(P_0)$ but our theory says that it could be enough.

\paragraph{Projection-Based Approximation.}
Given a “raw” (possibly uncentered) influence function estimate $\widetilde{D}_{\theta_{\text{targ}}}(O)$ at current $\theta_{\text{targ}}$ ---for example, influence function for an AIPW or IPTW-based estimator, or even $D^*_{P_{\theta_{\text{fix}},\,\theta_{\text{targ}}}}$ itself---we project $\widetilde{D}_{\theta_{\text{targ}}}(O)$ onto the partial derivatives of loss wrt $\theta_{\text{targ}}$:
\begin{align*}
    \alpha^*_{\theta_{\text{targ}}}
  \;=\;
  \arg\min_{\alpha}\!\sum_{i=1}^n 
    \Bigl[\widetilde{D}_{\theta_{\text{targ}}}(O_i)
          \;-\;\alpha^\top \nabla_{\theta_{\text{targ}}}\ell\bigl(\theta,O_i\bigr)\Bigr]^2
  \;+\;\lambda\,\|\alpha\|_2, 
\end{align*}
  
yielding 
\[
  D^*_{\mathrm{proj}}(O)=\hat{D}^{*,\text{para}}_{\theta_{\text{targ}}}
  \;=\;
  \alpha^*_{\theta_{\text{targ}}}\!{}^\top 
  \nabla_{\theta_{\text{targ}}}\ell\bigl(\theta,O\bigr).
\]

We omit the $\theta_{\text{targ}}$ in $D^*_{\mathrm{proj},\theta_{\text{targ}}}$ for simplicity. The regularization term $\lambda\,\|\alpha\|_2$ (or $\|\alpha\|_1$) helps stabilize the solution when the network is large and score gradients are correlated. The $\lambda$ can be chosen from cross validation or prespecified. 

This projection approach parallels regHAL's strategy~\cite{li2025regularized} 
of projecting influence functions onto score spaces, but generalizes from 
HAL's indicator basis functions to arbitrary neural network gradients 
$\nabla_{\theta_{\text{targ}}}\ell$.

\paragraph{Iterative Update with Targeting Gradient.}
Observe that $D^*_{\mathrm{proj}}(O)$ is linear in $\nabla_{\theta_{\text{targ}}}\ell$.  Thus, small steps of $\theta_{\text{targ}}$ in the direction $\alpha^*_{\theta_{\text{targ}}}$
define a \emph{locally least-favorable path} for debiasing/targeting in the submodel. So, we denote $\alpha^*_{\theta_{\text{targ}}}$ as our \emph{targeting gradient} at $\theta_{\text{targ}}$ for $\Psi(P_0)$. This eliminates the need to derive an analytic least-favorable submodel or compute a closed-form efficient influence function and we obtain the targeting direction entirely by solving a penalized regression on gradients.

In practice, we iteratively update $\theta_{\text{targ}}$:
\[
  \theta_{\text{targ}}^{(t+1)} 
  \;=\;
  \theta_{\text{targ}}^{(t)} 
  \;-\;\gamma_t\,\text{sign}\bigl[P_n(D^{*,(t)}_{\mathrm{proj}})\bigr]\,
          \alpha^*_{\theta_{\text{targ}}^{(t)}},
\]
or we use a simple 1D line search over $\gamma_t$. The use of $\text{sign}\bigl[P_n(D^{*,(t)}_{\mathrm{proj}})\bigr]$ is to guarantee we are minimizing the loss as we update. Convergence is checked by verifying 
\(
  \bigl|\,P_n[D^*_{\mathrm{proj}}]\,\bigr|\le\eta_n
\)
for the tolerance $\eta_n=\frac{sd(D^*_{\mathrm{proj}})}{\sqrt{n}logn}$ as suggested in \cite{rytgaard2024one} to guarantee the that $P_n(D^*_{\mathrm{proj}})=o_p(n^{-\frac{1}{2}})$, or upon reaching a max tolerance with no further loss improvement, which guarantees the convergence. Algorithm~\ref{alg:tda} below outlines these steps.

\begin{algorithm}[t!]
\caption{TDA Procedure (Projection-Based EIF Approximation)}
\label{alg:tda}
\begin{algorithmic}[1]
  \STATE {\bf Input:} Data $\{O_i\}_{i=1}^n$, initial $\theta_{\text{targ}}^{(0)}=\hat{\theta}_{\text{targ}}$, penalty $\lambda$, step size $\gamma_t$ or line search.
  \FOR{$t = 0,1,2,\dots$ \textbf{until convergence}}
     \STATE Compute $S_{\theta^{(t)}}(O_i) \;=\; \nabla_{\theta_{\text{targ}}^{(t)}}\ell\bigl(f_{\theta^{(t)}},O_i\bigr)$ for each $i$.
     \STATE Solve 
     \[
       \alpha^{(t)} \;=\; 
         \arg\min_{\alpha}\sum_i \bigl[\widetilde{D}(O_i) - \alpha^\top S_{\theta^{(t)}}(O_i)\bigr]^2 
         \;+\;\lambda\,\|\alpha\|_{1/\!2}.
     \]
     \STATE Define 
     \[
       D^{*,(t)}_{\mathrm{proj}}(O_i) 
       \;=\; 
       \alpha^{(t)\top}\,S_{\theta^{(t)}}(O_i).
     \]
     \STATE Update 
     \[
       \theta_{\text{targ}}^{(t+1)} 
         \;=\; 
         \theta_{\text{targ}}^{(t)} 
         \;-\;
         \gamma_t\,\text{sign}\!\Bigl[P_n\bigl(D^*_{\mathrm{proj}}\bigr)\Bigr]\,
                   \alpha^{(t)}.
     \]
     \STATE Check stopping: if $\bigl|P_n[D^*_{\mathrm{proj}}]\bigr|\le\eta_n$ or $t\!\ge\!T_{\max}$, break.
  \ENDFOR
  \STATE {\bf Output:} $\widehat{\theta}_{\text{targ}}^{(T)}$, and $\widehat{\Psi}=\Psi\bigl(\hat{\theta}_{\text{fix}},\widehat{\theta}_{\text{targ}}^{(T)}\bigr)$.
\end{algorithmic}
\end{algorithm}

\subsection{Multi-Parameter Targeting}
\label{sec:method:multi}

Many causal estimands are multi-dimensional, such as a \emph{survival curve} $\{S(t)\}_{t\in\mathcal{T}}$ or a vector of treatment effects under different doses.  Suppose each parameter $\Psi_k$ has an influence function $D_k(O)$.  In TDA, we simply project each $D_k$ separately and then \emph{combine} the resulting directions to create a universal canonical one-dimensional submodel for targeted minimum
loss-based estimation of a multidimensional target parameter as in \cite{van2016one}.  Concretely:
\[
  \alpha^*_k 
  \;=\;
  \arg\min_{\alpha}\!\sum_i \Bigl[
    D_k(O_i) \;-\;\alpha^\top S_{\theta}(O_i)\Bigr]^2 
    + \lambda\,\|\alpha\|_2,
  \quad
  d_k \;=\;\tfrac1n\sum_i\alpha_k^{*\top}\,S_{\theta}(O_i),
\]
then aggregate via
\[
  w_k \;=\; \frac{d_k}{\sqrt{\sum_{j=1}^K d_j^2}}, 
  \quad
  \alpha^* 
  \;=\;\sum_{k=1}^K w_k\,\alpha^*_k.
\]
A single step in direction $\alpha^*$ reduces the empirical mean of each $D_k$.  This ensures simultaneous debiasing without requiring separate iterative procedures for each parameter.

\vspace{0.5em}
\noindent \textbf{Remark.}  
Combining parameters is particularly powerful for \emph{functional} targets like survival curves, since we can preserve coherence across time points (such as monotone decreasing) while still pushing all marginal biases toward zero.  We next show that under standard regularity assumptions, these TDA updates inherit classical TMLE guarantees.  (Section~\ref{sec:theory}). 

\paragraph{Practical extensions (see Appendix \ref{sec:method-extensions}).}
We discuss two practical extensions—(i) adaptive sub-model expansion with
plateau-based selection, and (ii) TDA-Direct for special cases with
closed-form universal paths—in Appendix \ref{sec:method-extensions}.  
These options require no changes to the core algorithm yet broaden its
applicability in large architectures and simple ATE settings.

\section{Theoretical Guarantees}
\label{sec:theory}

Our consistency and efficiency results are a direct corollary of the \emph{Adaptive
Debiased Machine Learning (ADML)} framework of
van der Laan et al.~\cite{van2023adaptive} and its recent extension to \emph{smooth
M-estimands}~\cite{van2025automatic}.  We summarize the required assumptions
in section \ref{sec:theory:assump} for completeness and then specialize the general ADML
theorems to the neural sub-models generated by TDA.

\subsection{Oracle Submodel and Working Submodel}

Let $\mathcal{M}_{np}$ be the (nonparametric) model of all possible data distributions $P$ compatible with the observed variables.  We assume the true distribution $P_0\in \mathcal{M}_{np}$.  Our parameter of interest $\Psi:\mathcal{M}_{np} \to R $ is pathwise differentiable, so it admits an efficient influence function wrt $\mathcal{M}_{np}$ at $P$, denoted $D^*_{P}$. We are interested in target parameter $\Psi(P_0)$.

\paragraph{Oracle Submodel.}
While $\mathcal{M}_{np}$ is typically huge, suppose there exists a smaller ``oracle'' parametric submodel $M_0\subseteq \mathcal{M}_{np}$ in which $\Psi(P)$ can be estimated with negligible bias.  Concretely, for each $P$, let $\Pi_0P \in M_0$ be a suitable projection of $P$ onto $M_0$ (often a Kullback–Leibler or $L_2$ projection). Then we define the \emph{oracle parameter}:
\[
  \Psi_0(P) \;=\; \Psi(\,\Pi_0P\,),
  \quad
  \text{and specifically } 
  \Psi_0(P_0) = \Psi(\,\Pi_0P_0\,).
\]
In many cases, $\Pi_0P_0 = P_0$ itself if $P_0$ already lies in $M_0$, but we do not require exact coverage of $P_0$ if $M_0$ is slightly misspecified (i.e. $\Psi(P_0)-\Psi_0(P_0)$ is negligible).

\paragraph{Working Submodel via TDA.}
TDA defines a data-driven working submodel $M_n\subseteq \mathcal{M}_{np}$ by \emph{partially unfreezing} a subset of neural weights.  Formally, each distribution in $M_n$ corresponds to a neural net $f_{\theta_{\text{fix}}, \theta_{\text{targ}}}$, where $\theta_{\text{fix}}$ is held at its initial fit and $\theta_{\text{targ}}$ ranges over a $k_n$-dimensional space.  The induced \emph{working} parameter is
\[
  \Psi_n(P)
  \;=\;
  \Psi(\,\Pi_nP\,),
  \quad
  \Pi_n:\,\mathcal{M}_{np}\to M_n,
\]
an analogous projection onto the submodel $M_n$ (often a Kullback–Leibler or $L_2$ projection).  Thus, $\Psi_n(P_0)=\Psi(\Pi_nP_0)$ becomes our \emph{working target parameter}.

\subsection{High-Level Conditions and ADML Application}\label{sec:theory:assump}

Under ADML, the key idea is that if $M_n$ chosen by TDA is large enough to approximate the relevant part of $P_0$, and TDA solves the EIF of $\Psi(P)$ wrt $M_n$ (via the locally least-favorable path in weight space), then our final TDA plug-in estimator $\widehat{\Psi}_n=\Psi(P_{\textit{TDA}})$ inherits $\sqrt{n}$-consistency, semiparametric efficiency within $M_n$ for $\Psi(\Pi_nP_0)$, and can even achieve \emph{superefficiency} relative to $\mathcal{M}_{np}$ if $P_0$ indeed lies in a simpler submodel. More precisely, \cite{van2023adaptive} establishes conditions under which any \emph{adaptive} procedure that:

\begin{enumerate}[leftmargin=1em,itemsep=0pt]
\item Chooses a good submodel $M_n$ in a data-driven way (e.g., partial unfreezing, plateau selection),
\item Runs a TMLE-like update within $M_n$ to remove first-order bias for estimation of $\Psi(\Pi_nP_0)$,
\end{enumerate}

will enjoy the same asymptotic guarantees as if $M_0$ were known in advance.  Below, we restate the relevant ADML assumptions and theorems specialized to TDA.

\begin{enumerate} [leftmargin=1.5em,label=\textbf{(T\arabic*)}]
  \item\label{ass:A1A4}
        \emph{Regularity of the loss and $\Psi_0$}  
        (ADML Conditions A1–A4) including smoothness of loss function, pathwise differentiability of $\Psi_0$ at $P_0$. (Easily satisfied by most statistical problem set ups).
  \item\label{ass:B1}
        \emph{First-order expansion}  
        (ADML Condition B1):  
        the TDA estimator $\hat\Psi_n$ satisfies  
        $\hat\Psi_n=\Psi_n(P_0)+(P_n-P_0)D_{n,P_0}+o_p(n^{-1/2})$
        with $D_{n,P_0}$ the efficient influence function of
        $\Psi_n$ in $\mathcal{M}_n$ at $P_0$.  
        By construction TDA as a TMLE for $\Psi_n(P_0)$ in $M_n$ satisfies this, provided sample-splitting is used.
  \item\label{ass:B2B3}
        \emph{Gradient-coverage and empirical-process control}  
        (ADML Conditions B2–B3):  
        $\|D_{n,P_0}-D_{0,P_0}\|_{P_0}=o_p(1)$ and
        $(P_n-P_0)(D_{n,P_0}-D_{0,P_0})=o_p(n^{-1/2})$,
        where $D_{0,P_0}$ is the efficient influence function of
        $\Psi_0$ in $\mathcal{M}_0$ at $P_0$.  Practically in TDA, this amounts to requiring that the span of
        $\nabla_{\theta_{\mathrm{targ}}}\ell$ contains a good $L_2(P_0)$
        approximation to the oracle EIF $D_{0,P_0}$; in neural nets it can be checked by
        the residual norm of your projection step.
\end{enumerate}

\begin{enumerate}[leftmargin=1.5em,label=\textbf{(C\arabic*)}]
  \item\label{ass:C1}
        \emph{Projection of $P_0$ onto $M_n$ is nearly in $M_0$}  
        $\Psi\!\bigl(\Pi_nP_0\bigr)-\Psi\!\bigl(\Pi_0(\Pi_nP_0)\bigr)=o_p(n^{-1/2})$.
  \item\label{ass:C2}
        \emph{Oracle bias negligible}  
        $B_{n,0}+R_{n,0}=o_p(n^{-1/2})$, where these terms are defined as
        in Lemma 2 of ADML.
\end{enumerate}

\begin{theorem}[Asymptotic linearity for the working parameter]
\label{thm:AL-working}
Under \ref{ass:A1A4}–\ref{ass:B2B3}, TDA satisfies a first-order expansion, and empirical processes are controlled).  Then the TDA estimator $\widehat{\Psi}_n$ satisfies
\[
   \widehat{\Psi}_n 
   \;-\;
   \Psi_n(P_0)
   \;=\;
   (P_n - P_0)\,D_{0,P_0}
   \;+\;
   o_p(n^{-1/2}),
\]
where $D_{0,P_0}$ is the efficient influence function of the \emph{oracle} parameter $\Psi_0(P)$ wrt $\mathcal{M}_{np}$ at $P_0$.  In particular, $\widehat{\Psi}_n$ is $\sqrt{n}$-consistent and asymptotically normal for $\Psi_n(P_0)$.
\[
  \sqrt{n}\,\bigl\{\hat\Psi_n-\Psi_n(P_0)\bigr\}
  \;\;\xrightarrow{d}\;\;
  N\!\bigl(0,\;\Var_{P_0}\!\{D_{0,P_0}(O)\}\bigr).
\]
\end{theorem}

\begin{theorem}[Nonparametric efficiency and superefficiency]
\label{thm:efficiency}
Under additional conditions \ref{ass:C1}–\ref{ass:C2}, we have:
\begin{enumerate}[itemsep=0pt,topsep=0pt,label=(\roman*)]
  \item $\widehat{\Psi}_n$ is \emph{efficient} for the oracle parameter $\Psi_0(P_0)$, achieving the semiparametric variance bound in $M_0$,
  \[
  \sqrt{n}\,\bigl\{\hat\Psi_n-\Psi(\Pi_0P_0)\bigr\}
  \;\;\xrightarrow{d}\;\;
  N\!\bigl(0,\;\Var_{P_0}\!\{D_{0,P_0}(O)\}\bigr),
\]
  \item Consequently, since $\Psi(P_0)=\Psi(\Pi_0P_0)$, $\widehat{\Psi}_n$ is also \emph{superefficient} for the original $\Psi(P_0)$ in the full model $\mathcal{M}_{np}$, in that its asymptotic variance is no larger than the classical semiparametric bound in $\mathcal{M}_{np}$.
\[
  \sqrt{n}\,\bigl\{\hat\Psi_n-\Psi(P_0)\bigr\}
  \;\;\xrightarrow{d}\;\;
  N\!\bigl(0,\;\Var_{P_0}\!\{D_{0,P_0}(O)\}\bigr)
\]
\end{enumerate}
\end{theorem}

\paragraph{Perturbations and Local Alternatives.}
Under the same conditions, \cite[Theorem 6,7]{van2023adaptive} further establishes TDA remains \emph{regular and asymptotically linear} under local alternatives lying in $M_0$.  Particularly, no first-order efficiency loss occurs by adaptively selecting $M_n$ rather than fixing it a priori.

\paragraph{Multi-parameter targets.}
When $\Psi=(\Psi_1,\dots,\Psi_K)$ is vector-valued, stack the $K$
influence functions and apply the above results component-wise with help of the joint central limit theorem. Hence, TDA \emph{retains} the powerful inference guarantees of TMLE while allowing for flexible, high-dimensional architectures and data-dependent submodel choices—a combination well-suited to modern machine learning practice.

\section{Experiments}
\label{sec:experiments}

We benchmark \emph{Targeted Deep Architectures (TDA)} against both standard neural estimators and classical causal methods. Our primary metrics are (i) bias, (ii) variance, (iii) mean squared error (MSE), and (iv) confidence interval (CI) coverage in two canonical tasks:
\begin{enumerate}[label=(\alph*)]
    \item \textbf{Average Treatment Effect (ATE)} estimation on the IHDP dataset,
    \item \textbf{Survival curve} estimation under informative right censoring.
\end{enumerate}

\subsection{ATE Estimation on IHDP with DragonNet}
\label{sec:ate_ihdp}

\paragraph{Dataset and Setup.}
We use the Infant Health and Development Program (IHDP), a widely-adopted causal benchmark~\cite{hill2011bayesian,shalit2017estimating,shi2019adapting}, comprising 747 instances of low-birth-weight infants with 25 covariates ($X$), binary treatment ($A$), and semi-synthetic outcomes ($Y$). The true treatment effect is known, allowing an assessment of estimation bias and coverage. We split data 80\%–20\% for training and validation, then use the full sample for TDA updates. 

\paragraph{Implementation Details.}
We employ \emph{DragonNet}~\cite{shi2019adapting}, which includes a shared trunk (two layers, 64 units each, ELU activations) extracting latent features, a propensity-score head (one layer, 64 units, sigmoid output), and two outcome heads (treated, control), each one layer (64 units) predicting $\mu_1(X)$ or $\mu_0(X)$. We optimize with Adam (LR = 0.001), early-stopping on validation loss. TDA is applied after the initial training on full data. 

\paragraph{Compared Approaches.}
\begin{itemize}[leftmargin=1.5em]
    \item \textbf{Naïve Plug-in}: Directly uses $\hat{\mu}_1, \hat{\mu}_0$ for the ATE: 
    $
      \hat{\psi}_{\mathrm{plug}} 
      \;=\; \frac{1}{n}\!\sum_{i=1}^{n}\bigl(\hat{\mu}_1(X_i)-\hat{\mu}_0(X_i)\bigr).
    $
    \item \textbf{T-reg($\epsilon$)}: Targeted regularization through an added modification to the objective function as $\lambda[Y-Q-\epsilon(\frac{A}{g(X)}-\frac{1-A}{1-g(X)})]^2$ with $\lambda=0.01$ as in \cite{shi2019adapting}
    \item \textbf{Post-TMLE}: A classical post-hoc fluctuation adding a single $\varepsilon$-term to the outcomes, solving $P_n[D^*] \approx 0$ externally.
    \item \textbf{A-IPTW}: Augmented inverse-propensity weighting using $\hat{g}(X_i)$ and $\hat{\mu}_k(X_i)$, known to be doubly robust under correct nuisance fits.
    \item \textbf{TDA (Last Layer)}: Freezes all layers except the final outcome heads, using projection-based targeting with presepecified $\lambda=0.01$.
    \item \textbf{TDA (Full Layers)}: Unfreezes the entire outcome-head subnetwork (two layers), leaving the trunk and propensity head fixed, using projection-based targeting with prespecified $\lambda=0.01$.
\end{itemize}

\emph{Remark} The 95\% confidence intervals are computed as $\hat{\Psi}\pm 1.96 \sqrt{\frac{P_n(D^*_{\hat{u},\hat{g}})^2}{n}}$ for each method, where $D^*_{\hat{u},\hat{g}}$ is nonparametric EIF of the ATE at $\hat{u},\hat{g}$.

Full details of the data‑generating process and TDA implementation are provided in Appendix~\ref{app:ATE_IHDP}.

\paragraph{Results.}
After 1000 Monte Carlo replications (sample size $n=747$), all targeting methods, T-reg($\epsilon$) and A-IPTW reduce bias, variance and MSE significantly vs.\ naive plug-in (Table~\ref{table:ate}). Coverage rates improve for all adjustment methods, around 92\% for targeting methods and A-IPTW and 84\% for T-reg($\epsilon$)—much closer to nominal 95\% than the naive 74\%.  Overall, TDA’s targeted update within the neural architecture proves competitive with and even better in coverage than classical external TMLE while retaining inside the neural network architecture.

\begin{table}[h!]
\centering
\caption{Performance metrics for ATE estimation on IHDP (1000 replications).}
\label{table:ate}
\begin{tabular}{lcccc|cc}
\toprule
\multirow{2}{*}{Metric} & \multicolumn{1}{c}{Naive} & \multicolumn{1}{c}{T-reg} & \multicolumn{1}{c}{A-IPTW} & \multicolumn{1}{c}{Post-} & \multicolumn{1}{c}{TDA-} & \multicolumn{1}{c}{TDA-} \\
& \multicolumn{1}{c}{Plug-in} & \multicolumn{1}{c}{($\epsilon$)} & & \multicolumn{1}{c}{TMLE} & \multicolumn{1}{c}{Last} & \multicolumn{1}{c}{Full} \\
\midrule
Bias     & -0.178 & -0.110 & -0.139 & -0.135 & -0.137 & -0.122 \\
Variance & 0.165 & 0.046 & 0.063 & 0.060 & 0.091 & 0.084 \\
MSE      & 0.197 & 0.058 & 0.082 & 0.078 & 0.110 & 0.100 \\
Coverage & 0.735 & 0.840 & 0.915 & 0.915 & \textbf{0.920} & \textbf{0.924} \\
CI Width & 0.250 & 0.333 & 0.250 & 0.250 & 0.251 & 0.250 \\
\bottomrule
\end{tabular}
\end{table}

\subsection{Survival Curve Estimation Under Informative Censoring}
\label{sec:survival}

Based on the simulated right censored data $O=(X,\tilde{T}=\textit{min}(T,C),\Delta=1(T\leq C))$ according to the data generating process in Appendix \ref{sec:Survival set up}, we next estimate the marginal survival curve of $T$ in the presence of right censoring that is \emph{informative} (dependent on $X$). Our TDA approach naturally extends to multi-parameter functions by simultaneously targeting a fine grid of time points via a universal targeting gradient (Sec.~\ref{sec:method:multi}).

\paragraph{Setup and Comparison Methods.}
\begin{itemize}[leftmargin=1.5em]
    \item \textbf{Initial (Neural) Model}: Predicts $\log\lambda(t|X)$, the log of conditional hazard of $T$, with a multi-layer network that is trained via partial-likelihood or Poisson-likelihood approximation.
    \item \textbf{Kaplan-Meier}: The classical nonparametric survival estimator. In principle, it does not account for informative censoring but is widely used for baseline comparison.
    \item \textbf{TDA (Survival)}: Our universal targeting approach, adjusting the final layer to align the network’s predicted curve at equally spaced $50$ points with influence-function projections that accounts for censoring and the hazard form with $\lambda=1e-5$. 
\end{itemize}

\emph{Remark.} We omit a direct comparison with a single-step or universal-fluctuation TMLE for the entire survival function since it quickly becomes analytically cumbersome and computationally heavy in high dimensions~\cite{rytgaard2024one,laan2003unified}. Adapting classical TMLE in the neural netwrok set up for large time-discretized survival data would require a specialized submodel for each time point or a complicated integral-hazard universal path, making it impractical for broad usage. TDA’s embedded approach is precisely designed to circumvent this complexity.

Full details for TDA implementation are provided in Appendix~\ref{sec:Survival set up}.

\paragraph{Metrics} 
We evaluate survival estimation accuracy across 50 targeted time points, reporting means and standard deviations (±) across replications. Time-averaged MSE quantifies overall error by averaging $(S_{\text{est}}(t) - S_{\text{true}}(t))^2$ across time points. Absolute bias measures systematic error through $|S_{\text{est}}(t) - S_{\text{true}}(t)|$. Variance represents the variability of estimates across replications at each time point, then averaged. For the initial and TDA, coverage indicates the percentage of times the true survival falls within model-generated confidence intervals using estimated submodel EIC at each time point, then averaged. For Kaplan-Meier, we use the greenwood formula couterpart.

\paragraph{Results.}
We ran 500 replications with $n=1{,}000$ samples, time-varying hazard, and 30\% informative censoring. Table~\ref{table:survival} shows TDA reduces time-averaged MSE by 47.06\% versus the initial neural network and 44.49\% versus Kaplan-Meier. Absolute bias decreases by 25.89\% compared to the initial model and 24.72\% versus Kaplan-Meier. Despite slightly higher variance, TDA's overall accuracy is substantially better. TDA's confidence intervals achieve approximately 91\% coverage. More visualization are available in the appendix 

\begin{table}[h!]
\centering
\caption{Survival curve estimation under informative censoring (500 replications).}
\label{table:survival}
\begin{tabular}{lccc}
\toprule
Metric & Initial Model & Kaplan-Meier & TDA (Survival) \\
\midrule
MSE     & 0.0029 (±0.0031)  & 0.0028 (±0.0026)  & \textbf{0.0015 (±0.0014)} \\
Abs. Bias  & 0.0413 (±0.0225)  & 0.0407 (±0.0204)  & \textbf{0.0306 (±0.0149)} \\
Variance       & 0.0027  & 0.0028   & 0.0014  \\
\midrule
Coverage (\%) & 75.51 & 55.84 & 90.98 \\
CI Width       &  0.1608  & 0.0755   & 0.4773  \\
\bottomrule
\end{tabular}
\end{table}

\section{Discussion}
\label{sec:discussion}

\noindent\textbf{Summary and Key Advantages.}
\emph{Targeted Deep Architectures (TDA)} embeds TMLE’s low‐dimensional ``$\varepsilon$‐fluctuation’’ directly into a neural network’s parameters by freezing most weights and updating only a small “targeting” subset. This yields a scalable, architecture‐agnostic procedure that \emph{debiases} neural predictions and produces valid \(\sqrt{n}\)-confidence intervals. Crucially, TDA handles \emph{multiple} or \emph{functional} targets—like entire survival curves—by combining multiple influence directions into a single universal update.

\noindent\textbf{Limitations and Open Problems.}
(\textit{i}) \emph{Non-convex convergence:} Although iterative local updates are typically stable, formal global guarantees are lacking.  
(\textit{ii}) \emph{Non-i.i.d.\ data:} TDA relies on standard cross‐fitting for i.i.d.\ samples; dependent structures (time-series, panel data) need specialized influence expansions.  
(\textit{iii}) \emph{Memory overhead:} Large architectures may pose challenges when computing high‐dimensional gradient projections; approximate factorization or low‐rank gradient representations could mitigate cost.  
(\textit{iv}) \emph{Overfitting checks:} Partial unfreezing avoids extreme variance inflation, but synergy with neural regularization techniques (e.g.\ dropout, weight decay) warrants further study.

\noindent\textbf{Complementary streams.} We see TDA as complementary to two parallel streams:
(i)~the \emph{regHAL--TMLE} line of work~\cite{li2025regularized}, which realizes similar projection ideas for HAL-based working models via explicit ridge/lasso fluctuations. Like TDA, regHAL addresses stability issues arising from collinear basis functions by projecting influence functions onto score spaces with regularization. TDA extends this projection-based philosophy from HAL's indicator basis functions to arbitrary neural network gradients, demonstrating that these regularization principles apply broadly across modern deep architectures.
(ii)~the broader \emph{Adaptive Debiased Machine Learning} framework \cite{van2023adaptive}, whose theory underpins our sub-model selection and efficiency guarantees. Within ADML, autoDML~\cite{van2025automatic} provides an elegant approach for smooth functionals of M-estimands when the Hessian operator is well-conditioned. TDA complements autoDML by extending ADML's applicability to settings where Hessian-based approaches face challenges—particularly in overparameterized neural networks where Hessian operators are often singular or computationally prohibitive. Moreover, TDA's gradient-projection approach handles any pathwise differentiable parameter with a known influence function, not just those arising from M-estimand functionals. This includes: (i) quantile treatment effects where the parameter isn't defined through smooth loss minimization, (ii) survival parameters where closed-form efficient influence functions may not exist but IPTW-based influence functions are available, and (iii) complex functionals where practitioners have derived influence functions through other means. By requiring only gradients rather than Hessian operators, TDA brings the benefits of ADML—efficiency, double robustness, and data-adaptive model selection—to the challenging setting of modern deep learning architectures while maintaining the computational tractability demonstrated by regHAL's projection-based approach.

\noindent\textbf{Future Directions.}
One path is integrating the TDA targeting gradient \emph{during} training, blending predictive and influence-based objectives. Another is advanced submodel selection (e.g.\ second-order methods) to identify which layers to unfreeze. Extending TDA to dynamic or panel data would also broaden its scope.

\noindent\textbf{Conclusion.}
By localizing TMLE’s fluctuation within a small subset of network weights, TDA achieves double robustness, efficiency, and multi-parameter scalability without cumbersome post-hoc steps. Across benchmark ATE and survival tasks, TDA significantly improves inference quality—highlighting its promise for flexible, rigorous causal analysis in modern deep learning.

\begin{ack}

This material is based upon work supported by the National Science Foundation Graduate Research Fellowship Program under Grant No. DGE 2146752. Any opinions, findings, and conclusions or recommendations expressed in this material are those of the author(s) and do not necessarily reflect the views of the National Science Foundation.

\end{ack}

\medskip
\bibliographystyle{plain}
\bibliography{references}

\newpage
\appendix

\section{Appendix: Experimental Details}
\label{app}
This appendix provides additional information on dataset characteristics, implementation details, hyperparameter choices, and supplementary results for the experiments described in Section~\ref{sec:experiments} of the main text.
\subsection{ATE Experiments on IHDP}
\label{app:ATE_IHDP}
\paragraph{Dataset Characteristics.}
The IHDP dataset combines covariates from a real randomized experiment studying educational interventions for premature infants with semi-synthetic outcomes. It contains 747 units with:
\begin{itemize}[leftmargin=15pt]
    \item 25 covariates (6 continuous, 19 binary) measuring child and maternal characteristics
    \item Binary treatment indicator (educational intervention)
    \item Synthetically generated outcomes following the "Response Surface B" setting from \cite{hill2011bayesian}, where the treatment effect varies with covariates
    \item Mean treatment effect approximately 4.0 (varying slightly across samples)
\end{itemize}
We analyze multiple samples (denoted "replications" in the code) to ensure our findings are robust to the particular synthetic outcome realization. Each replication uses a different subset of units from the full dataset.

\paragraph{Data Splitting.}
For each replication, we implement a three-fold data splitting strategy:
\begin{itemize}[leftmargin=15pt]
    \item \textbf{Training set (80\%)}: Used for initial model fitting
    \item \textbf{Validation set (20\%)}: Used for early stopping and hyperparameter tuning
    \item \textbf{Targeting set (100\%)}: Full data harnessed for the targeting procedure
\end{itemize}
This splitting approach mitigates potential overfitting in both the initial model training and the subsequent targeting step, providing more reliable estimates of generalization performance. For each fold, we ensure balanced representation of treatment and control groups.
\paragraph{DragonNet Implementation Details.}
\begin{itemize}[leftmargin=15pt]
\item \textbf{Architecture}: We implement DragonNet in PyTorch with:
    \begin{itemize}
        \item 2 trunk layers (64 units each, ELU activation)
        \item 1 propensity head layer (64 units, ELU activation → sigmoid output)
        \item 1-layer outcome heads for both treatment and control (64 units, ELU activation)
    \end{itemize}
\item \textbf{Training}: Adam optimizer with learning rate 0.001, weight decay 1e-5, batch size 128
\item \textbf{Early stopping}: Based on validation loss with patience of 10 epochs
\item \textbf{Loss function}: Binary cross-entropy for propensity + MSE for outcome prediction

\end{itemize}

\paragraph{TDA Implementation.}
For the targeting procedure, we implement two variants:
\begin{itemize}[leftmargin=15pt]
\item \textbf{TDA (Last Layer)}: We freeze all network parameters except the final linear layers of the outcome heads (one each for $\mu_0$ and $\mu_1$). This restricts targeting to a small number of parameters (approximately 65 weights per outcome head).
\item \textbf{TDA (Full Layers)}: We allow targeting to update both layers in each outcome head, keeping the trunk and propensity head fixed. This increases the targeting dimension to approximately 4160 parameters per outcome head.
\end{itemize}
The targeting procedure uses the following steps:
\begin{enumerate}[leftmargin=15pt]
\item \textbf{Hybrid projection system}: We compute $\delta_i$ residuals (following the EIF) for each sample, and build a design matrix of MSE gradients with respect to targeted parameters.
\item \textbf{Regularized least squares}: We solve $(G^TG + \lambda I)\Delta w = G^T r$ with $\lambda = 0.01$ for robust projection.
\item \textbf{Parameter updates}: We update parameters iteratively with line search.
\item \textbf{Adaptive early stopping}: When mean residual is sufficiently small or convergence plateaus.
\end{enumerate}
\paragraph{Comparison Methods.}
\begin{itemize}[leftmargin=15pt]
\item \textbf{Initial (Plug-in)} uses $\hat{\psi}_{ATE} = \frac{1}{n}\sum_{i=1}^{n}(\hat{\mu}_1(X_i) - \hat{\mu}_0(X_i))$
\item \textbf{TMLE} adds a single $\varepsilon$-fluctuation step to the outcome predictions using the clever covariate $H(X) = \frac{A}{\hat{p}(X)} - \frac{1-A}{1-\hat{p}(X)}$
\item \textbf{Double-Robust (DR)} ATE estimates are computed for all methods as:
$$
\hat{\psi}_{DR} = \frac{1}{n}\sum_{i=1}^{n}\left[ \frac{A_i}{\hat{p}(X_i)}(Y_i - \hat{\mu}_1(X_i)) - \frac{1-A_i}{1-\hat{p}(X_i)}(Y_i - \hat{\mu}_0(X_i))\right]
$$
\end{itemize}
\paragraph{Confidence Intervals.}
95\% confidence intervals are constructed as $\hat{\psi} \pm 1.96 \times \hat{\sigma}/\sqrt{n}$, where $\hat{\sigma}^2$ is the sample variance of the estimated efficient influence function values.

\subsection{Survival Curve Experiments}
\label{sec:Survival set up}
\paragraph{Data Generation.}
We simulate survival data from a complex generative process with:
\begin{itemize}[leftmargin=15pt]
\item \textbf{Covariates}: $X \sim \mathcal{N}(0, \Sigma)$ with $\Sigma_{ij} = 0.3^{|i-j|}$ (moderate correlation)
\item \textbf{True event times}: Generated from a complex hazard model:
\[
\lambda(t|X) = \lambda_0(t) \cdot \exp\left(\beta_{tv}(t)^TX_{1:3} + \sum_{j=4}^6\exp(X_j\beta_j) + \sum_{j=7}^{10}\sum_{k>j}^{10}X_jX_k\beta_j\beta_k\right)
\]
where $\lambda_0(t) = \frac{\alpha}{\eta}\left(\frac{t}{\eta}\right)^{\alpha-1}$ is a Weibull baseline with $\alpha=1.5$ (increasing hazard) and $\eta=10$ (scale), and $\beta_{tv}(t) = \beta_{1:3}\sqrt{t}$ are time-varying coefficients.
\item \textbf{Informative censoring}: Censoring times $C$ generated from:
\[
C \sim \text{Weibull}\left(\alpha_c, \eta_c \cdot \exp\left(\gamma^T X + \sum_{j=2}^4|X_j|^{1.5}\right)\right)
\]
where coefficients $\gamma$ are chosen to create correlation between censoring and event mechanisms, with $\gamma_j$ positive for $j \leq 5$ (increasing censoring risk) and negative for $j > 5$ (decreasing censoring risk).
\item \textbf{Observed data}: $\tilde{T} = \min(T, C)$ and $\Delta = \mathbf{1}{(T \leq C)}$, with approximately 30\% censoring rate. And also covariates $X$.
\end{itemize}
The time-varying effects and interactions create complex survival patterns that violate proportional hazards assumptions. The informative censoring creates additional challenges for standard survival methods.
\paragraph{Neural Hazard Model.}
We implement a neural network for conditional hazard estimation:
\begin{itemize}[leftmargin=15pt]
\item \textbf{Architecture}:
\begin{itemize}
\item Input layer: covariates $X$ concatenated with time $t$
\item 3 hidden layers with dimensions [64, 32, 16]
\item ReLU activations and dropout (rate 0.2)
\item Final linear layer to predict $\log\lambda(t|X)$
\end{itemize}
\item \textbf{Training}:
\begin{itemize}
\item Poisson log-likelihood loss: $\ell(\lambda, \Delta, \tau) = \tau\lambda - \Delta\log(\lambda\tau)$
\item where $\tau$ is time-at-risk for each interval
\item Adam optimizer with learning rate 0.001
\item Early stopping based on validation loss
\item Mini-batch size 64
\end{itemize}
\end{itemize}
For model evaluation, we compute survival functions by numerical integration of the predicted hazards:
\[
\hat{S}(t|X) = \exp\left(-\int_0^t \hat{\lambda}(u|X)du\right)
\]
\paragraph{Projection-Based Targeting for Survival.}
Our TDA approach for survival analysis has several distinct components:
\begin{enumerate}[leftmargin=15pt]
\item \textbf{Score matrix computation}: We construct a score matrix representing model gradients with respect to the final layer parameters, capturing the relationship between model parameters and the EIF.
\item \textbf{Censoring adjustment}: We fit a separate neural network to model the conditional censoring process $G(t|X)$, and use these estimates to weight observations in the targeting procedure, addressing the informative censoring.
\item \textbf{Universal targeting}: For $m$ time points of interest ${t_1, \ldots, t_m}$, we:
\begin{itemize}
\item Compute influence functions for each time point $D_{t_j}(O)$
\item Project each onto the score space: $\alpha^j = \arg\min\alpha \sum_i (D_{t_j}(O_i) - \alpha^T S{\theta}(O_i))^2 + \lambda|\alpha|_1$
\item Compute empirical means: $d_j = \frac{1}{n}\sum_i D_{t_j}(O_i)$
\item Form a weighted update direction: $\alpha^* = \sum_{j=1}^m w_j \alpha^j$ with $w_j = d_j / ||d||_2$
\item Update model parameters: $\theta_{targ} := \theta_{targ} + \varepsilon\alpha^*$
\end{itemize}
\item \textbf{Iterative updates}: Continue updating until convergence criteria are met or maximum iterations reached.
\end{enumerate}
This approach allows us to simultaneously target the entire survival curve, maintaining coherence and smoothness across time points.
\paragraph{Confidence Interval Construction.}
For the neural network models, we compute 95\% confidence intervals using the efficient influence function approximation:
\begin{itemize}[leftmargin=15pt]
\item Project the EIF for each time point onto the score space using regularized regression
\item Compute the empirical standard error of the projected influence function
\item Form confidence intervals as $\hat{S}(t) \pm 1.96 \cdot \hat{\sigma}_t / \sqrt{n}$
\end{itemize}
For the Kaplan-Meier estimator, we use the standard Greenwood formula for variance estimation.
\paragraph{Evaluation Metrics.}
We evaluate performance using:
\begin{itemize}[leftmargin=15pt]
\item \textbf{Integrated MSE}: $\int_0^{\tau} (\hat{S}(t) - S_0(t))^2 dt$, approximated by numerical integration
\item \textbf{Absolute bias}: $\int_0^{\tau} |\hat{S}(t) - S_0(t)| dt$
\item \textbf{Pointwise coverage}: Percentage of time points where $S_0(t) \in [\hat{S}(t) \pm 1.96 \cdot \hat{\sigma}_t / \sqrt{n}]$
\item \textbf{Average CI width}: Mean width of confidence intervals across time points
\end{itemize}

\subsubsection{Visualization of survival–curve accuracy}
\label{sec:survival_viz}

To give an intuitive account of how the estimators behave over time, we aggregate the
results from independent Monte-Carlo replications (Section~\ref{sec:Survival set up}) and
display them in two complementary plots.

\begin{figure}[ht]
  \centering
  \includegraphics[width=0.9\linewidth]{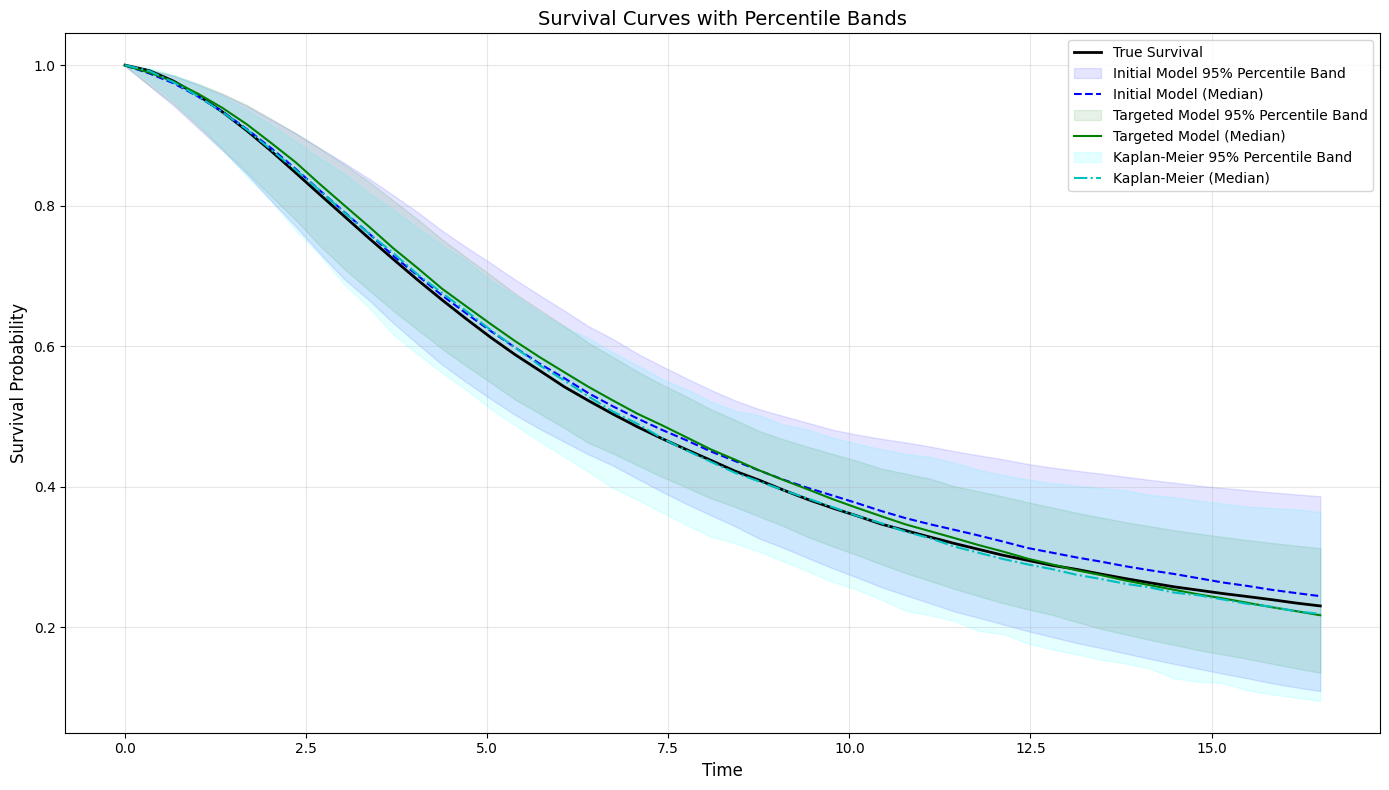}
  \caption{\small
    \textbf{Median survival curves with $95\%$ percentile bands.}
    Solid curves show the median survival estimate across replications
    for the initial model (blue), the targeted model (green), and the
    Kaplan–Meier estimator (cyan).  Shaded ribbons depict the corresponding
    $2.5$th–$97.5$th percentiles.  The black line is the
    ground-truth survival function used in the simulation.}
  \label{fig:survival_band}
\end{figure}

\paragraph{Uncertainty bands for the survival function.}
For each of the $50$ evaluation times $t\in[0,16]$ we collect the estimated survival
probabilities, compute their empirical median, and derive a $95\%$ percentile band
($2.5$th–$97.5$th percentiles) across replications.
Figure~\ref{fig:survival_band} juxtaposes the
\emph{true} survival curve (black) with the median trajectories
of (i)~the initial neural network fits, (ii)~our TDA fits, and
(iii)~the non-parametric Kaplan–Meier baseline.
Shaded ribbons of decreasing opacity illustrate the associated
uncertainty bands.  Visually, the TDA estimator tracks the ground truth
more closely than its competitors, while maintaining tighter bands for most of the time horizon.

\begin{figure}[ht]
  \centering
  \includegraphics[width=0.9\linewidth]{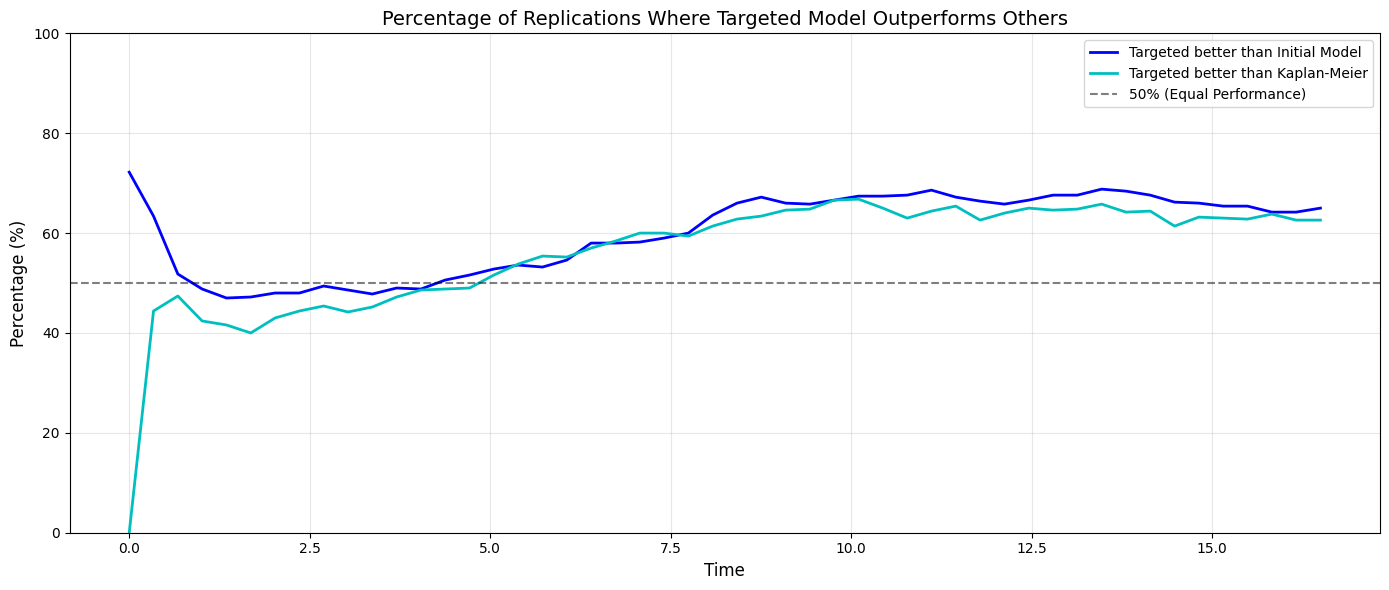}
  \caption{\small
    \textbf{How often does targeting help?}
    Percentage of replications in which the targeted estimator has a smaller
    absolute error than the initial model (blue) and the Kaplan–Meier estimator
    (teal) at each time point.  The horizontal dashed line indicates parity
    ($50\%$); values above the line mean the targeted estimator is more accurate
    in the majority of runs.}
  \label{fig:outperform}
\end{figure}

\paragraph{How often does the targeted estimator win?}
Figure~\ref{fig:outperform} answers this question directly.
At every evaluation time we record whether the TDA’s absolute error
with respect to the truth is \emph{smaller} than that of the initial neural network (blue)
or the Kaplan–Meier estimator (teal).  The figure reports the proportion of
replications for which the targeted model is superior.  The dashed grey line
marks the $50\%$ threshold: values above that line mean the targeted estimator
outperforms the competitor in the majority of runs.  Apart from the very early
time points (where all estimators are nearly unbiased), the targeted estimator
wins in roughly $60\!$–$70\%$ of replications—underscoring its practical benefit.

\section{TDA Extensions}
\label{sec:method-extensions}

\subsection{Adaptive Expansion of the Submodel}
\label{sec:method:tmle_req}

A central principle of classical TMLE is that its ``$\varepsilon$-parametric submodel'' must be sufficiently rich to capture the necessary fluctuation that drives $P_n[D^*(\theta)]=0$. In TDA, this translates to having:

\[
\left\|\text{EIF}_{\text{relevant}} - \text{Proj}_{\mathrm{span}\{\nabla_{\theta_{\text{targ}}}\,\ell(\theta)\}}(\text{EIF}_{\text{relevant}})\right\|_{P_0} = o_p(1)
\]
ensuring we can move along a least-favorable path within the network’s weight space. If we freeze too many layers, the final-layer submodel may be overly restrictive, making it impossible to eliminate first-order bias. Conversely, if the submodel is large enough to represent the EIF direction, TDA inherits classical TMLE guarantees such as double-robustness (if one nuisance function is correct) and local semiparametric efficiency.

\paragraph{From Last-Layer to Partial Unfreezing.}
A common practical choice is to freeze all but the \emph{final} layer, which typically yields enough flexibility for many tasks (e.g.\ average treatment effect). However, in settings with richer structure (multiple treatments, survival analysis, complex interactions), the final layer alone may not capture the necessary adjustments.  In such cases, TDA can \emph{selectively} unfreeze additional blocks of weights (layers, channels) that strongly influence the target parameter:
\begin{itemize}[leftmargin=1em]
\item \textbf{Gradient-based selection:} For each block of parameters $\theta_{j}$, approximate 
\(\bigl\|\nabla_{\theta_j}\Psi(\hat{\theta})\bigr\|\)
and unfreeze those blocks with large magnitude.  Intuitively, blocks whose gradients have minimal impact on $\Psi$ can remain frozen, while high-impact blocks become part of $\theta_{\text{targ}}$.
\item \textbf{Second-order/Hessian criteria:} Identify blocks that yield large expected reduction in $P_n[D^*(\theta)]$ per unit norm, e.g., using approximate Hessian or curvature information.  Only blocks with the most ``leverage'' are unfrozen.
\end{itemize}
This systematic approach yields a nested sequence of candidate submodels $M_1\subset M_2\subset\dots\subset M_K$ of increasing dimension. One might unfreeze all weights---but this risks overfitting in finite samples.

\paragraph{Plateau-Based Submodel Selection.}
To avoid both underfitting (too few unfrozen parameters) and overfitting (too large a submodel), we can adapt \emph{plateau selection} (a common strategy in sieve TMLE \cite{davies2014sieve}).  Specifically:
\begin{enumerate}[leftmargin=1em,itemsep=0pt]
\item For each candidate submodel $M_k$, run TDA to obtain $\widehat{\Psi}_k$ plus an approximate standard error $\widehat{se}_k$.
\item Form a confidence interval $\bigl[\widehat{\Psi}_k - 1.96\,\widehat{se}_k,\;\widehat{\Psi}_k + 1.96\,\widehat{se}_k\bigr]$.
\item Compare the \emph{lower} or \emph{upper} interval bound (depending on whether $\widehat{\Psi}_k$ is trending upward or downward).  Once the bound plateaus or expands again, stop.
\end{enumerate}
Intuitively, if a smaller $M_k$ already drives $P_n[D^*]$ near zero, further expansions waste degrees of freedom and may inflate variance unnecessarily.  If $M_k$ is insufficient, we see persistent bias, prompting us to expand.  By halting once the CI width no longer meaningfully grows, TDA remains aligned with standard TMLE theory while balancing bias and variance in weight-space submodels.

\paragraph{Relation to Classical TMLE Expansions.}
In classical TMLE, one might add successive fluctuation terms $e_1, e_2,\dots$ to reduce bias.  Here, each fluctuation is effectively realized by \emph{expanding} the parameter set in the neural net.  If too many fluctuation terms are introduced (or too large a submodel is unfrozen), we can overfit.  TDA’s partial unfreezing plus a plateau rule provides a data-driven way to regulate this tradeoff: start with a smaller submodel and progressively unfreeze more weights only if necessary.  
\medskip

\noindent In summary, partial unfreezing and adaptive submodel selection are natural analogs to adding multiple ``$\varepsilon$-fluctuations'' in classical TMLE.  They ensure TDA can handle scenarios where a single-layer submodel is insufficient, while providing a data-driven mechanism (the plateau selector) to avoid overfitting in high-dimensional neural architectures.

\subsection{TDA-Direct:  A Streamlined Targeting Step for some Special Cases (e.g., the ATE)}
\label{sec:tda_direct}

The general TDA algorithm (Sections~\ref{sec:method}–\ref{sec:method:multi})
projects an estimated influence function onto the span of the score vectors
$\nabla_{\theta_{\mathrm{targ}}}\!\ell$ in order to construct an (implicit)
least–favourable path.
For many causal parameters, however, a closed-form \emph{linear universal}
least-favourable submodel is already known.  
A canonical example is the average treatment effect (ATE), whose update is
driven by the \emph{clever covariate}
\[
  H(A,X)\;=\;\frac{A}{g(X)}-\frac{1-A}{1-g(X)} ,
\]
where $g(x)=\Pr(A=1\mid X=x)$.  
In such settings we can bypass the projection step entirely and
embed the fluctuation directly in the \emph{last layer} of the outcome network
$Q(a,x)=\E[Y\mid A=a,X=x]$.  
We denote this simplified variant \textbf{TDA-Direct}.

\paragraph{Algorithm.}
Assume the final layer of $Q$ is linear, i.e.\ 
$
  Q_{\theta}(a,x)=\sum_{j=1}^{M}\theta^{\mathrm{last}}_{j}\,
  \phi_{j}^{(\mathrm{early})}(a,x)
$
with fixed early-layer features $\phi_{j}^{(\mathrm{early})}$.
\begin{enumerate}[leftmargin=1.65em,label=(\arabic*)]
  \item \textbf{Initial fits.}  
        Train a network for $Q$ and a separate model for the propensity
        score~$g$ (possibly another head of the same network).
  \item \textbf{Compute clever covariate.}  
        Evaluate $\hat{H}_i = \hat{H}(A_i,X_i)$ for all observations.
  \item \textbf{Linear regression in weight space.}  
        Freeze all weights \emph{except} $\theta^{\mathrm{last}}$ and solve
        the regression
        \[
          \hat{\alpha}
            =\arg\min_{\alpha\in\mathbb{R}^{M}}
            \sum_{i=1}^{n}
            \Bigl(\hat{H}_i-\alpha^{\!\top}\phi^{(\mathrm{early})}(A_i,X_i)
            \Bigr)^{2}.
        \]
        The solution $\hat{\alpha}$ is our \emph{targeting gradient} and we hope to overfit $\hat{H}_i$ as much as we can.
        
  \item \textbf{One-step update.}  
        Replace the last-layer weights by
        $
          \theta^{\mathrm{last}} \leftarrow
          \theta^{\mathrm{last}}+\varepsilon\,\hat{\alpha},
        $
        where $\varepsilon$ is found by a one-dimensional line search that
        minimises the empirical loss for~$Q$.
        The resulting network $Q^{\star}$ defines the TDA-Direct estimate.
\end{enumerate}

\paragraph{Proof sketch.}
Because the fluctuation is linear in $\hat{H}$, the universal least-favourable
path for $Q$ can be written
\[
  Q_{\theta^{(0)}}(\varepsilon)
  \;=\;
  Q_{\theta^{(0)}} \;+\; \varepsilon\,\hat{H}.
\]
If $\hat{H}$ can be expressed in the span of the final-layer features,
i.e.\ $\hat{H}\approx\sum_{j}\alpha^{\star}_j\phi^{(\mathrm{early})}_j$, then
\[
  Q_{\theta^{(0)}}(\varepsilon)
  \;\approx\;
  \sum_{j=1}^{M}
  \bigl(\theta^{\mathrm{last}(0)}_{j}+\varepsilon\alpha^{\star}_j\bigr)\,
  \phi^{(\mathrm{early})}_j,
\]
so updating $\theta^{\mathrm{last}}$ by $\varepsilon\alpha^{\star}$ reproduces
the least-favourable path within network‐parameter space.  Choosing
$\varepsilon$ to drive the empirical mean of the EIF to $o_{\!p}(n^{-1/2})$
therefore yields the same first-order bias removal guaranteed by TMLE.

\paragraph{Remarks.}
\begin{enumerate}[leftmargin=1.4em,label=(\roman*)]
  \item \textbf{Direct modelling of $H$.}  
        If $H$ is obtained from a separate regression rather than via
        $\hat{g}$, Step~(2) still applies unchanged; Step (1) omits the
        propensity-score fit.
  \item \textbf{Expressiveness.}  
        The key requirement is that $\hat{H}$ lies ( or at least approximately) in the span
        of the final-layer features.  When this is doubtful, one can add
        \emph{skip connections} from deeper hidden units to the output so that
        the final linear layer spans a richer function class while remaining
        linear in~$\theta^{\mathrm{last}}$ during targeting.
\end{enumerate}

Such special cases with an universal least favorable path that is linear in clever co-variates which only depends on the orthogonal nuisance parameters are prevalent such as the average treatment effect in the longitudinal case. \textbf{TDA-Direct} thus serves as a more simple and fast alternative to the general TDA discussed in the main paper.

\section{Computation and Code Availability}
\label{app}
All experiments were implemented in Python 3.9.6 and executed on machines with Apple M4 Pro chips. Code to reproduce all experiments is available on GitHub at https://github.com/yiberkeley/TDA. The implementation includes:
\begin{itemize}[leftmargin=15pt]
\item \emph{ATE\_TDA / dragonnet\_TDA.py}: Implementation for the ATE experiments on IHDP data based on dragonnet.
\item \emph{ATE\_TDA / run\_dragonnet\_TDA.py}: Parallelize the dragonnet\_TDA.py
\item \emph{ATE\_TDA / summary\_dragonnet\_TDA.ipynb}: Evaluation metrics and confidence interval computations
\item
\emph{Survival\_TDA / HazardNN\_TDA.ipynb}: Implementation for survival analysis with neural hazards
\item \emph{Survival\_TDA / summary\_hazardNN\_TDA.ipynb}: Evaluation metrics and confidence interval computations
\end{itemize}
Runtime for a single replication of the IHDP experiment is approximately 2 minutes on an Apple M4 Pro chip, and 5-7 minutes for a survival analysis replication.

Our code builds on public implementations of DragonNet and neural hazard models, with significant modifications to implement the novel TDA methods. We also provide scripts to generate the simulated survival data.

\end{document}